\documentclass[twoside,leqno,twocolumn]{article}

\usepackage[letterpaper]{geometry}

\usepackage{ltexpprt}

\usepackage{cite}
\usepackage{amsmath,amssymb,amsfonts}
\usepackage{algorithmic}
\usepackage{graphicx}
\usepackage{textcomp}
\usepackage{xcolor}
\usepackage{multirow}
\newtheorem{definition}{Definition}

\def\BibTeX{{\rm B\kern-.05em{\sc i\kern-.025em b}\kern-.08em
    T\kern-.1667em\lower.7ex\hbox{E}\kern-.125emX}}
    
\begin{document}

\title{Optimize Cash Collection: Use Machine learning to Predicting Invoice Payment}

\author{Ana Paula Appel\thanks{IBM Research - apappel@br.ibm.com} \and Victor Oliveira\thanks{UFABC - Work during internship - vt.victoroliveira@gmail.com}  \and Bruno Lima\thanks{IBM Research - blima@br.ibm.com}  \and Gabriel Louzada Malfatti\thanks{IBM Research - gabriel.malfatti@ibm.com} \and Vagner Figueredo de Santana\thanks{IBM Research - vagsant@br.ibm.com}  \and Rogerio de Paula\thanks{IBM Research - ropaula@br.ibm.com} }
    
\date{}

\maketitle
%

\fancyfoot[R]{\scriptsize{Copyright \textcopyright\ 2020 by SIAM\\
Unauthorized reproduction of this article is prohibited}}

\begin{abstract}

Predicting invoice payment is valuable in multiple industries and supports decision-making processes in most financial workflows. 
However, the challenge in this realm involves dealing with complex data and the lack of data related to decisions-making processes not registered in the accounts receivable system.
This work presents a prototype developed as a solution devised during a partnership with a multinational bank to support collectors in predicting invoices payment.
The proposed prototype reached up to 77\% of accuracy, which improved the prioritization of customers and supported the daily work of collectors.
With the presented results, one expects to support researchers dealing with the problem of invoice payment prediction to get insights and examples of how to tackle issues present in real data.
Keywords: machine learning, account receivables, feature engineering, payment, finance
\end{abstract}

%

\section{Introduction}

%

The invoice-to-cash process involves various steps, from invoice creation to customer's debt (payment) settlement or reconciliation. One key step of this process is the collection of accounts receivable. Accounts receivable (AR) refers to the invoices issued by a company for products or services already delivered but not yet paid for by its customers. Properly managing AR is a core accounting activity and concern of any company, pertaining to its cash-flow. 

Despite the recent widespread use and adoption of information technologies, particularly, machine learning techniques, across domain and industry applications, there are still companies that manage internal processes in ways they did in the past, with paper and pencil. 


In this work, we present a study case carried out in partnership with a multinational bank (hereafter also referred to as client).
It sought for innovative ways to proactively identify overdue ARs with high probability of being paid such that its managers and executives could take appropriate actions (such as, reaching out to those customers and collecting those ARs). As an international bank, it operates in multiple countries; however, this project focuses exclusively on its customers based in Latin America. 

The collection activity is performed by analysts, also known as collectors. The collectors are responsible for charging bank's customers (hereafter referred to as customers) and in turn improving these customers' experience relative to the payment processes. In the bank, each collector deals with approximately 100 customers and these customers are allocated according to collectors' seniority level, i.e., senior collectors are responsible for bigger accounts and/or contracts. In collecting ARs (i.e. the activity of charging customers), collectors receive daily a list prioritizing customers to contact. This list takes into account a customer's debt based on all of its overdue invoices. Nonetheless, it ignores the customer's payment behavior, such as, whether this customer regularly pays on time or not. Usually, customers are contacted at a fixed schedule before the due dates, irrespective of whether a particular customer regularly pays its invoices on time or not. Neither does it differentiate a recurrent from a sporadic payment behavior, such as, an occasional financial problem faced by a customer. However, in the end, all of this detailed information about customers' payment behaviors lays in individual collectors' minds, being utilized just in an ad hoc manner. 

\begin{figure}[hbt]
    \centering
    \includegraphics[width=0.46\textwidth]{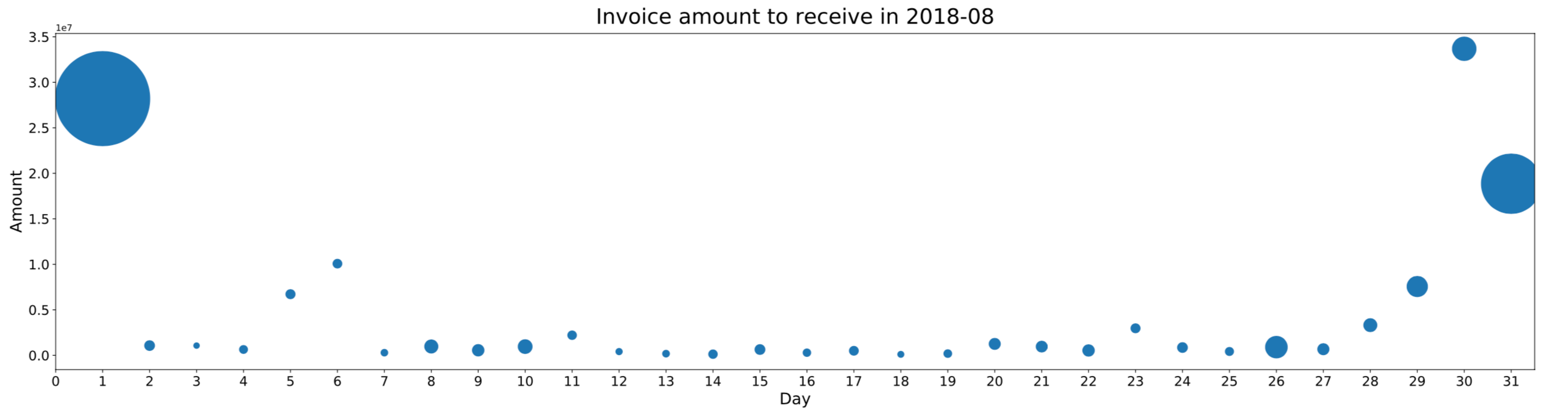}
    \caption{Invoices to be receive over one month, distributed over the payment data. Each point could have one or more invoices.}
    \label{fig:montlycollect} 
\end{figure}

Figure \ref{fig:montlycollect} represents what collectors face every day, which is dealing with a large number of invoices from several clients in a month. Each bubble is a set of invoice to be receive in a day, the size of  bubble means how many invoices are in that day and these invoices could be from one or more clients. The position of the bubbles are the amount of money to be receive that day. Looking at this Figure how the collector should decide who he/she will call first? Which client is more likely to pay late? how they keep track of clients behavior? So, what is the obvious choice is to go for the large amount of money, since in the end of the month the collectors performance will be evaluate by the amount of money that he/she recovery. But we know that the largest amount of money is not necessary from the higher risk client. 
Hence, predicting invoice payment (i.e., the invoices most likely to be paid next) can be a solution to a better allocation of resources and a better cash flow estimation, essential to achieving financial stability. 

In our research, we assert that by providing insights as to how to prioritize contacting clients based on the probability of late payment helps collectors make more effective and efficient decisions. The objective is to help them make more assertive and timely decision by means of focusing their collection actions on invoices that would have a great financial return at the same time that would be most likely to respond to collectors solicitation (i.e. being paid). To this end, the system would provide a personalized, ranked list of customers to whom to contact, taking into account a collector's own list of customers and their payment behavior.

\begin{figure}[hbt]
    \centering
    \includegraphics[width=0.46\textwidth]{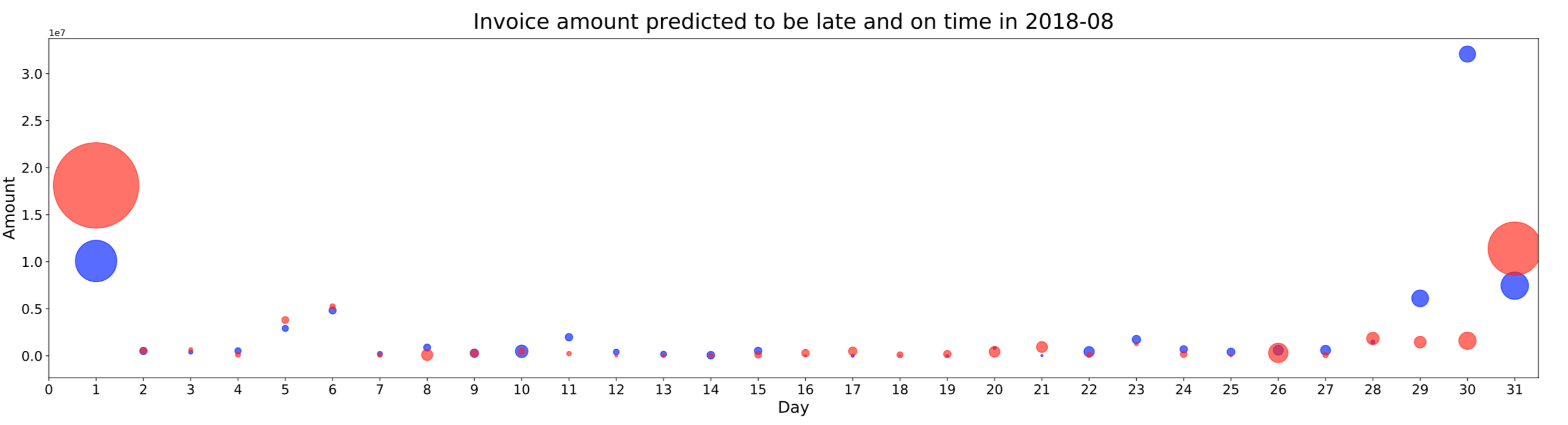}
    \caption{Invoices to be receive over one month, distributed over the payment data. Each point could have one or more invoices.}
    \label{fig:montlycollectpredict1} 
\end{figure}

Figure \ref{fig:montlycollectpredict1} shows when we apply the model prediction developed for us how we can make collector's life better. Knowing that some of the invoices will be payed on time, collector can only follow to see if the invoice will be really payed and focus on clients that are predict to be late. 

Predictive modeling approaches are widely used in a number of related domains, such as, credit management and tax collection \cite{abe2010optimizing}.
The problem of predicting invoice payment has been traditionally tackled using statistical survival analysis methods, such as, the proportional hazards method \cite{Lee:2013}. 
Survival analysis is a statistic method for analyzing the expected duration of time until one or more events happen, such as death in biological organisms and failure in mechanical systems.

Dirick et al. \cite{Dirick2017} tested several survival analysis techniques in credit data from Belgian and Great Britain financial institutions. Survival analyses were also experimented in modeling consumer credit risk \cite{cao2009modelling, cheong2018customer, rychnovsky2018survival}. These works focus on predicting ``when'' an event may occur rather than ``whether'' it may occur or not. This aligns with our interest in analyzing time to an event; thus, survival analysis approach is a reasonable technique for tackling the problem at hand. 

Smirnov \cite{smirnov2016modelling} concluded that \textit{Random Survival Forests} model, which additionally uses historical payment behavior of debtors, performs better in ranking payment times of late invoices than traditional \textit{Cox Proportional Hazards} model. Although the \textit{proportional hazards} model is the most frequently used model for survival analysis, it still has a number of drawbacks, such as the baseline hazard function is uniform and proportional across the entire population, as explained by \cite{baesens2005neural}.

Invoice payment prediction could also be modeled as a classification problem, but there is just a small body of work that addresses this problem. One of the few works that investigate this is Zeng~\cite{zeng2008using}, where the authors formulate the problem as traditional supervised classification and apply existing classifiers to it. They divided the clients into four different classes, such as, on time, 1-30 days, 31-60 days, and +60 days. These classes are usually related to AR process and the counter measures for addressing late invoices. Similarly, Bailey et al.~\cite{bailey1999} analyze several strategies for prioritizing collection calls and propose to use predictive modeling based on binary logistic regression and discriminative analysis to determine which customers to hand over to an outside collections agency for further collection processing.

Tater et al. \cite{tater2018prediction} propose a different approach to the problem and instead of predicting invoice in accounts receivable, they focus on accounts payable, working on invoices that were already delayed. Similarly, Younes \cite{younes2013framework} focuses on accounts payable case and attempts to address the problem of invoice processing time, understanding the overdue invoices and the impact of delays in the invoice processing.
Abe et al. \cite{abe2010optimizing} in addition propose a new approach for optimally managing the tax and, more generally, debt collections processes at financial institutions.


The prototype herein described aimed at devising and developing a tool employing state-of-the-art Artificial Intelligence (AI) algorithms and techniques for creating a ranked list of (potential) overdue accounts for each collector, based on different criteria, such as, the highest probability of payment in the short-term, payment behavior patterns, and the like. 
It thus aimed at optimizing collectors' actions and thus improving the payment rate of these accounts.

The key contribution of our project are: 
\begin{itemize}
\item The use of machine learning to predict with high accuracy the status of invoices (late or on time), allowing the bank to estimate better how much money will be delayed in cash and work pro-actively to avoid late payments; 
\item The use of historical and temporal features to improve model's accuracy and an extensive comparison of models to use the one that fits best;
\item An effective new way for ranking customers to be prioritized by collectors not only taking into account the volume of money, but also the probability of being late;
\end{itemize}

The paper is organized as follows. Section \ref{sec:problem} defines the problem formally. Section \ref{sec:data} characterizes the data set used in the work and the ETL (Extract, Transform, Load) process. 
Section \ref{sec:experiments} shows the modeling approach applied in the problem and the results obtained so far. Section \ref{sec:ranking} presents the prioritizing list proposed to client.  
Finally, Section \ref{sec:conclusion} discusses outcomes and concludes our work. 

\section{Problem Definition} \label{sec:problem}

In AR collection, the ability of monitoring and collecting payments foments the prediction of payment behavior. Firms often use various types of metrics to measure the performance of the collection process, for instance, average number of days overdue. Particularly in our case the client is interested mainly in knowing the probability that an invoice will be paid late or on time to then be able to better prioritize the collection.  

The problem of predicting an invoice payment is a typical classification problem using supervised learning \cite{James:2014} where, given the original client dataset, we need to extract invoices' features to be able to characterize each invoice with respect to labeled classes, building then a machine learning model to perform classification of new invoices.

In a more formal way, we can define our problem as follows:

\begin{definition}
Let $\mathcal{M} = {I,Y}$ be a set of pairs of invoices and their respective classes. Element $M_m$ is represented by the pair $\langle I_m, Y_m \rangle$, with $I_m$ represented as a set of features $A = {a_1, a_2,..., a_n}$, and the class $Y_m$ having a binary value representing either a \textit{``late''} or \textit{``on time''} state for $I_m$.
\end{definition}

In order to prepare the dataset for the model training, we defined class $Y_m$ for each invoice $I_m$. The definition of class $Y_m$ as \textit{``on time''} or \textit{``late''} is done as follows: 

\begin{equation}
\small
  Y_m=
    \begin{cases}
      on\;time, & \text{if}\ payment\;at\;most\;5\;days\;from\;due\;date \\
      late, & \text{otherwise}
    \end{cases}
\end{equation}

Thus, an invoice is considered overdue if the payment occurred in more than 5 days from due date. The main reason for considering this time window is the time required to processing payment in the client system. This interval was elicited during one of the meetings we had with client's subject matter experts (SMEs) to understand the problem, processes, and work flow of the collection activity.

\section{Data Source and ETL} \label{sec:data}

The dataset received has 175,552 invoices from 8 countries from Latin America, 3,725 customers ranging from August 2017 to June 2019. The invoices distribution by country are presented in Table \ref{tab:invoicecountry}. Since we have some countries with low representativeness and after analyzing the payment behaviour of the countries with high representativeness in the dataset, we develop only one model instead of one for each country.

\begin{table}[htb]
\centering
\begin{tabular}{ |l|r|r| } 
\hline
Country & \# invoices & \# customers \\ 
\hline
Argentina & 337 & 33 \\ 
\hline
Brazil & 46,262 & 1,265 \\ 
\hline
Chile & 21,565 & 339 \\ 
\hline
Colombia & 27,960 & 614 \\ 
\hline
Ecuador & 20 & 1 \\ 
\hline
Mexico & 53,010 & 844 \\ 
\hline
Peru & 25,884 & 604 \\ 
\hline
Uruguay & 514 & 25 \\ 
\hline
\end{tabular}
    \caption{Data distribution for Latin America showed by country. Some countries are very under represented in the data, thus we aimed at building one model for all countries instead of a particular model for each country.}
    \label{tab:invoicecountry}
\end{table}

The dataset we studied only contains information about payments, i.e., only about invoice. For instance, invoice value, country code, customer number, etc. One of biggest problem of our data is that our client has no relevant information about customers as industry sector, balance sheets, etc.  We had only a number to identify customers uniquely without any other information. 
Since the scope and time of the project does not include gather information about the client and due to privacy constrains we accept the challenge of work only with invoice information.






One of the biggest challenges were how to transform the poor invoice information in relevant features to build a machine learning model to predict payment. 
As traditional real data is, we found a large number of missing values and wrong information in several data fields that would be important for machine learning model. One of the main reasons is that all the data available comes form legacy system spread over the world from other sectors and business units that we are working on consolidates the data. This is one of the reasons that some of costumer data is not available.   
In order to extract the most valuable features from the data and get it right we perform several discussions with client's SMEs and collectors, looking at previous literature on AR, and thorough a detail examination of the data. 

We start with traditional invoice-level features, such as: invoice amount, credit rating, etc. 
In order to enrich our model with more significant information than only invoice-level features as supra-cited, we performed a feature extraction to build the late invoice payment model.

We used historical data to create aggregate features that could bring more meaning to our set of invoice-level features. As reported in \cite{zeng2008using}, the use of aggregate features increase significantly the amount of information about payment.
However, some of the features recommended in \cite{zeng2008using} did not work in our case, specially due to some data related issues. One example are the ratios that, in our case, because of a lot of missing information, increase the number of null values in the data. Also, some features, as category that specify if an invoice was under dispute or not, had a lot of consistency issues due to manually entered information. 

On the other hand, we incorporated some features related to the recent payments in order to capture customer behavior. Based on our carefully analyze on the data we observed that recent payments influence more in the payment behaviour than older payments.
These features are whether the customer paid each of its last three invoices, percentage of paid invoices, payment frequency, number of contracts related to each invoice, and standard deviation of invoices late and outstanding. Table \ref{tab:features} shows the constructed features as well as the respective descriptions.

\begin{table*}[htb]
\centering
\begin{tabular}{|p{7cm}|p{9cm}|} 
\hline
Features generated & Description \\ 
\hline 
\textit{paid invoice} & Value indicating whether the last invoice was paid or not; where 1 means paid, 0 means not paid, and -1 indicates null value (possible first time customer). \\ \hline
\textit{total paid invoices} & Number of paid invoices prior to the creation date of a new invoice of a customer. \\ \hline
\textit{sum amount paid invoices} &
The sum of the base amount from all the paid invoices prior to a new invoice for a customer.\\ \hline
\textit{total invoices late} &
Number of invoices which were paid late prior to the creation date of a new invoice of a customer. \\ \hline
\textit{sum amount late invoices} &
The sum of the base amount from all the paid invoices which were late prior to a new invoice for a customer. \\ \hline
\textit{total outstanding invoices} &
Number of the outstanding invoices prior to the creation date of a new invoice of a customer. \\ \hline
\textit{total outstanding late} &
Number of the outstanding invoices which were late prior to the creation date of a new invoice of a customer. \\ \hline
\textit{sum total outstanding} &
The sum of the base amount from all the outstanding invoices prior to a new invoice for a customer. \\ \hline
\textit{sum late outstanding} &
The sum of the base amount from all the outstanding invoices which were late prior to a new invoice for a customer. \\ \hline
\textit{average days late} &
Average days late of all paid invoices that were late prior to a new invoice for a customer. \\ \hline
\textit{average days outstanding late} &
Average days late of all outstanding invoices that were late prior to a new invoice for a customer \\ \hline
\textit{standard deviation invoices late} &
Standard deviation of all invoices that were paid late. \\ \hline
\textit{standard deviation invoices outstanding late} &
Standard deviation of days late of all outstanding invoices that were late prior to a new invoice for a customer. \\ \hline
\textit{payment frequency difference} &
Amount of times the customer did a payment. Intention here is to identify customers that payed more invoices. \\ \hline
\end{tabular}
    \caption{Feature extracted in order to build a model to predict invoice payment. We used historical features and invoice level features combined to bring more significant information to the model. The use of invoice level features only resulted in a poor model. In addition, we did not use any feature related to the customer due to lack of structured information and privacy issues as well.}
    \label{tab:features}
\end{table*}

The next step in our ETL process was to handle missing values. We had to do this for invoices that did not have too much historical information accordingly our features.
One of our case was that we had a null value for sum of total invoices, we just replaced it by zero. However, in a few cases it was necessary taking in consideration the feature's representation. For average days late, we could not fill with zeros, as it is an indicative of good payment behavior. In these cases, we used mean as a replacement value for missing ones. 

With the dataset cleaned up and properly set, we could start to work in the best way to split in train, test and validation sets. 
To prevent data leakage, that is, create an accurate model to make predictions on new data, unseen during training, we split our dataset regarding time. Thus, split in train, test, and validation sets based on time of invoice creation. For training, we considered data ranging from August 2017 to July 2018, validation set from August 2018 to November 2018, and test from December 2018 to June 2019. Table \ref{tab:modelrange} presents in details the amount of invoices that we had in each of our dataset. As we can see, the distribution of invoices between late and on time is a little unbalanced towards late invoices, except that in the test set we have the same distribution between late and ontime invoices.  

\begin{table*}[htb]
\large
\begin{center}
\begin{tabular}{ |l|r|r|r|r|c| } 
\hline
Dataset & \# invoices & Late & On Time & Baseline & Period \\ 
\hline
Train & 115,503 & 68.94\% & 31.06\% & 68.94\% & 2017-08/2018-07 \\ 
\hline
Validation & 24,891 & 61.70\% & 38.30\% & 61.70\% & 2018-08/2018-11 \\ 
\hline
Test & 35,158 &  52.58\% & 47.42\% & 52.58\% & 2018-12/2019-06 \\ 
\hline
\end{tabular}
\end{center}
    \caption{Data distribution for training, validation and test sets. The classes are balanced and our baseline is close to 65\%. We split the data using time, since we cannot use future data to make predictions. We use around 70\% of available data to train the model and the other 30\% we split in test and validation data.}
    \label{tab:modelrange}
\end{table*}


\section{Modeling Approaches and Results} \label{sec:experiments}

As explained before, our problem was defined as a binary classification problem to predict if either an invoice will be payed on time or late. Although we stated the problem as predicting classes, a wide range of models return probabilities instead of just labels. This is crucial in order to do a prioritization list and rank customers with higher chances of default. Also, since the model was planned to be deployed in a client that lately will need to retrain and update the model, it is important that we use a powerful model in terms of scalability, handle missing values, and would be easy do understand the results and retrain so the non machine learning experts could have a sense about what is going on with the data.  

We tested our data with five different classification methods: Naive Bayes, Logistic Regression, k-Nearest Neighbors, Random Forest~\cite{Breiman:2001}, Gradient Boosted Decision Trees~\cite{FRIEDMAN2002367}. 



Most of features came from historical data, for example, \textit{sum amount late invoices}, \textit{total invoices late} and so on. In order to calculate these features for an invoice, we needed to define a period of time that we will consider to look back. This period is different from our trained dataset that defines which invoices we will consider. To define the best range of time to look back to calculate the features, we created a parameter that we call \texttt{window size}.
In short, window size will be the number of months prior to an invoice that we will consider to calculate our features values. 
But, why not just extract features based on all data set and avoid an extra step? 
The problem is that as time passes, the statistical distribution of the features also changes and reduces model's accuracy. In the machine learning and predictive analytics realm this is known as concept drift. Therefore, it is necessary to work with boundaries and to focus on getting information from the most recent past, representing the most recent customer behavior. 

In order to create a more robust model and to make sure that we are using the correct time range, we created 11 datasets with $w$ ranging from 2 to 12 months to performer tests and see how many months do we need to consider.

Table \ref{tab:results} summarizes the results for each model and all the $w$ considered. 
As we observed, XGB and Random Forest achieved the best accuracy with $w=2$, while Logist Regression and Naive Bayes with $w=2$. K-NN achieved its best at $w=27$. This means that, using more data than we really need to calculate our features could degrade the model instead of making it better. 

Actually as we can see, a small number of months works better than a large one. This is specially because the concept drift in the data. In our dataset, the customers are performing better over the months, that is, they are paying invoices less late than in the starting point of the dataset. Thus, using all the data to calculate historical features would insert a bias in the dataset.

\begin{table*}[htb]
\centering
\begin{tabular}{|l|r|r|r|r|r|} \hline 
 \multirow{2}{*}{Window Size} & \multicolumn{5}{c|}{Classifiers} \\ \cline{2-6}
 & XGB & Random Forest & Logistic Regression & Naive Bayes & k-NN \\ \hline  
2-months &	\textbf{79.75\%} & \textbf{79.05\%} &	74.91\% & 72.00\% & 75.40\%  \\ \hline 
3-months &	77.32\% & 76.54\% & 75.82\% & 74.53\% & 75.56\% \\ \hline 
4-months &	76.78\% & 76.45\% &	76.12\% & 74.54\% & 75.62\% \\ \hline 
5-months &	76.95\% & 76.75\% & \textbf{76.43\%} & \textbf{75.08\%} & 75.36\% \\ \hline 
6-months & 76.45\% & 76.26\%  & 76.58\% & 74.99\% & 76.17\% \\ \hline
7-months &	76.63\% & 76.76\% & 76.01\% & 74.95\% &	\textbf{76.84\%} \\ \hline 
8-months &	76.47\% & 76.95\% & 76.29\% & 74.78\% &	75.65\% \\ \hline 
9-months &	76.82\% & 76.48\% &	76.16\% & 74.81\% &	75.79\% \\ \hline 
10-months & 76.90\% & 76.74\% & 76.23\% & 74.53\% & 75.17\% \\ \hline
11-months & 76.93\% & 76.88\% & 75.98\% & 74.45\% & 75.62\% \\ \hline
12-months & 77.37\% & 76.96\% & 76.40\% & 74.11\% & 76.45\% \\ \hline 
\end{tabular}
\caption{We present the accuracy results for all the data set generated from $w=2,..,12$ with all five classifiers that we tested: Naive Bayes, Logistic Regression, k-Nearest Neighbors, Random Forest, and Gradient Boosted Decision Trees. We highlight the best accuracy and we can see that with 2 months ($w=2$) we achieved the best performance in XGB and Rando Forest, the other classifiers had worse performance and need more historical data ($w=5$ and $w=7$ for k-NN). This result shows that using all the available data is not always the best option. Specially because the concept drift we observed lead to use less data ($w=5$).}
\label{tab:results}
\end{table*}

Despite the fact of the best accuracy was achieved with two months, in our feature engineering we have a feature that look at three last payments and we know that our data is very susceptible to data drift and other external variances (as politics, economy, etc). Base on this, we decide to use 3 months as $w$ so we also want avoid over-fit the model for the data. All the bellow experiments were done with $w=3$.

\begin{figure}[hbt]
    \centering
    \includegraphics[width=0.47\textwidth]{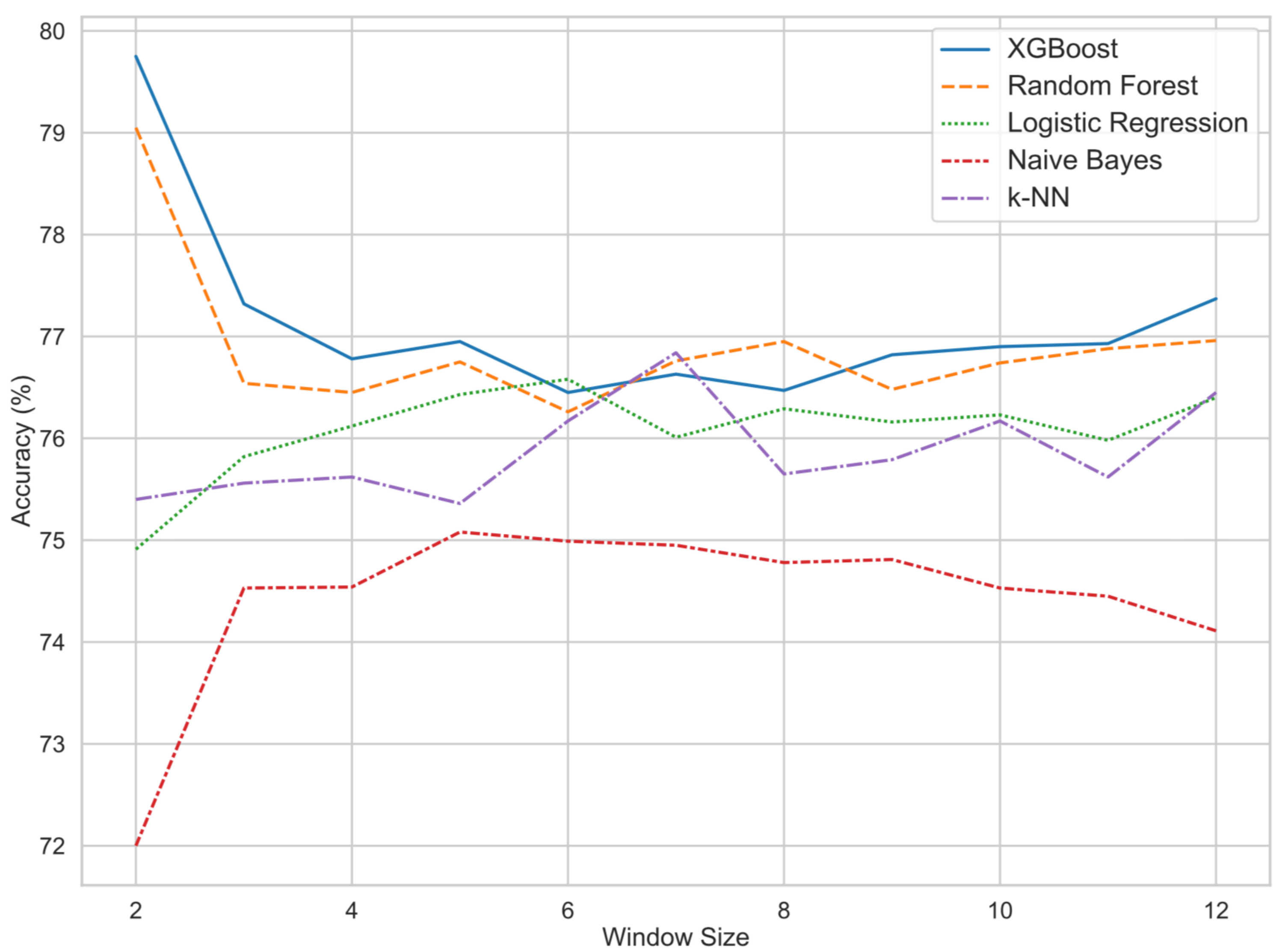}
    \caption{Plot showing accuracy for all tested models with all the $w$ generated. The ensemble methods are the ones with high accuracy in all $w$, being $w=5$ the best result. Naive Bayes is the one that scores worst, but it achieved its best performance with $w=2$. We see that more we accumulate our historical features, the worst the models perform. }
    \label{fig:modelacuracy}
\end{figure}

Figure \ref{fig:modelacuracy} shows the results for all models and $w$ values. Compared to the baseline score (59.39\%), we can see that the models perform significantly better, specially the ensemble learning models (Random Forest and Gradient Boosting) are the ones with higher accuracy. 



Giving that our model returns a probability score rather than a label indicating whether it is from class late or on time, we can rank the targets and use it to plot the Receive Operating Characteristic (ROC) curves in order to compare models. They depict the performance of a classifier comparing the True Positive Rate vs False Positive Rate. True Positive Rate, also known as recall metric, indicates that given a true class, for example, class late, what is the percentage of samples classified as late compared to the ground-truth, that is, the total number of instances from class late. On the other side, False Positive Rate is the percentage of falsely reported positives out of the ground-truth negatives (class on time). Intuitively, the ROC curves will give a guidance to understand how well a model is performing based on a ranking, that is, if we have a high lift point on the curve, which demonstrates that invoices with late labels have higher probability score of being late (as expected).

\begin{figure}[h]
    \includegraphics[width=0.47\textwidth]{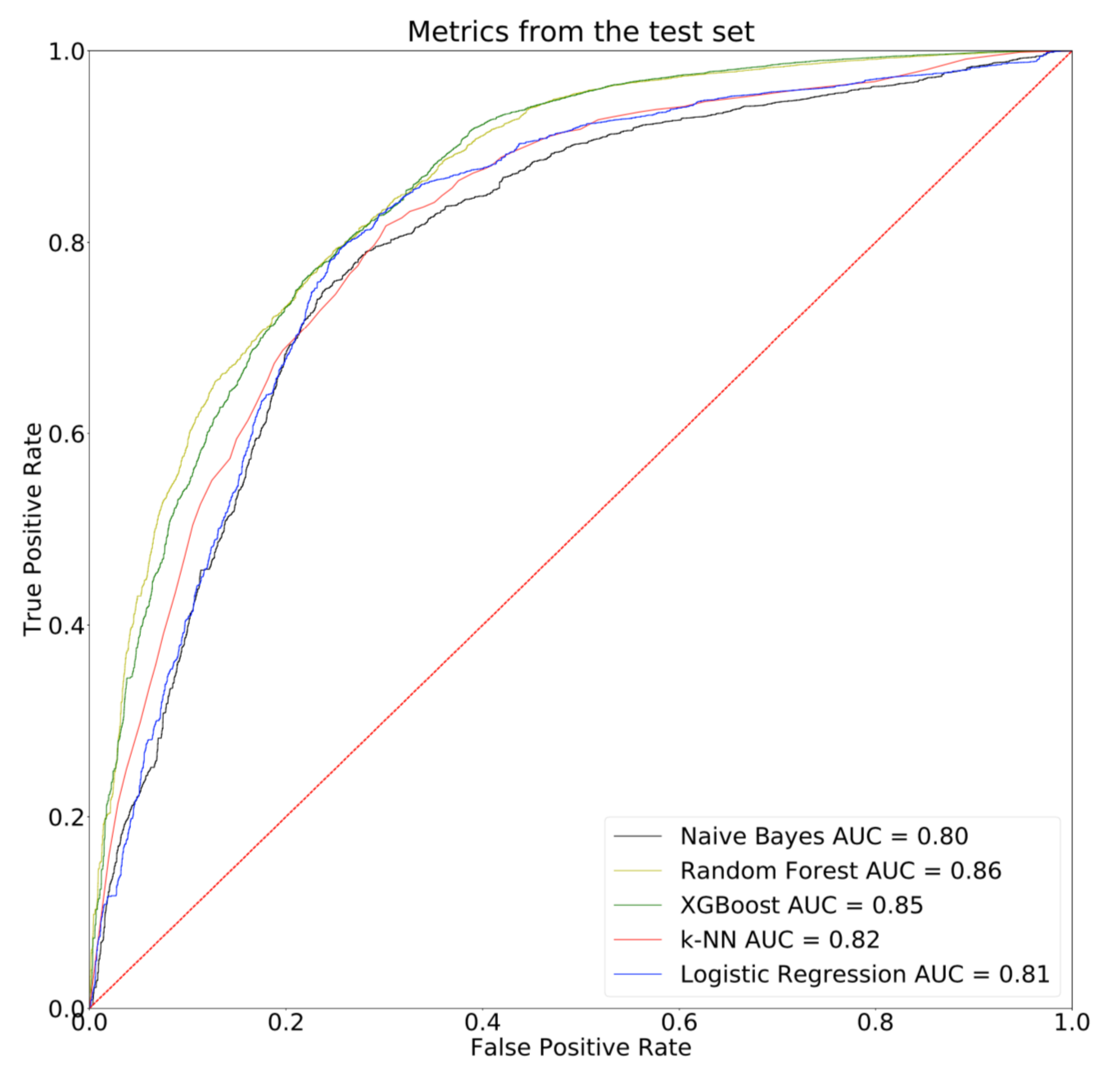}
    \caption{ROC Curve of all five methods we tested in our test dataset with $w=3$.}
    \label{fig:rocv}
\end{figure}

In order to measure not only graphically but also quantitatively, we used the area under the curve metric (AUC) as well. AUC is a metric that calculates the area under our ROC curve, i.e., it as a way of calculating the lift point explained above. The Figure \ref{fig:rocv} shows the ROC curves for each model and the corresponding AUC score for test and validation.

Clearly, the best models are the Random Forest and Gradient Boosting. Therefore, we developed the predictive invoices label system based on an ensemble approach using both models.

Another important point to be evaluate is the amount of data used to train, validation and test the model. We know that our data suffer from a large concept drift, that is, over the years the clients are performing better which means that they are paying less late and more on time. Also, we know that payment data is very sensitive to external factors, as seasonality, country economy, politic subjects, etc. We can add to this the question that we use a hold-out option since we do not want to use future data to predict past data. To analyze this, we create several snapshot of the data varying period of data. We generate four snapshot as presented in Table \ref{tab:results2}, always moving the train to more recent data. 

\begin{figure}[h]
    \includegraphics[width=0.47\textwidth]{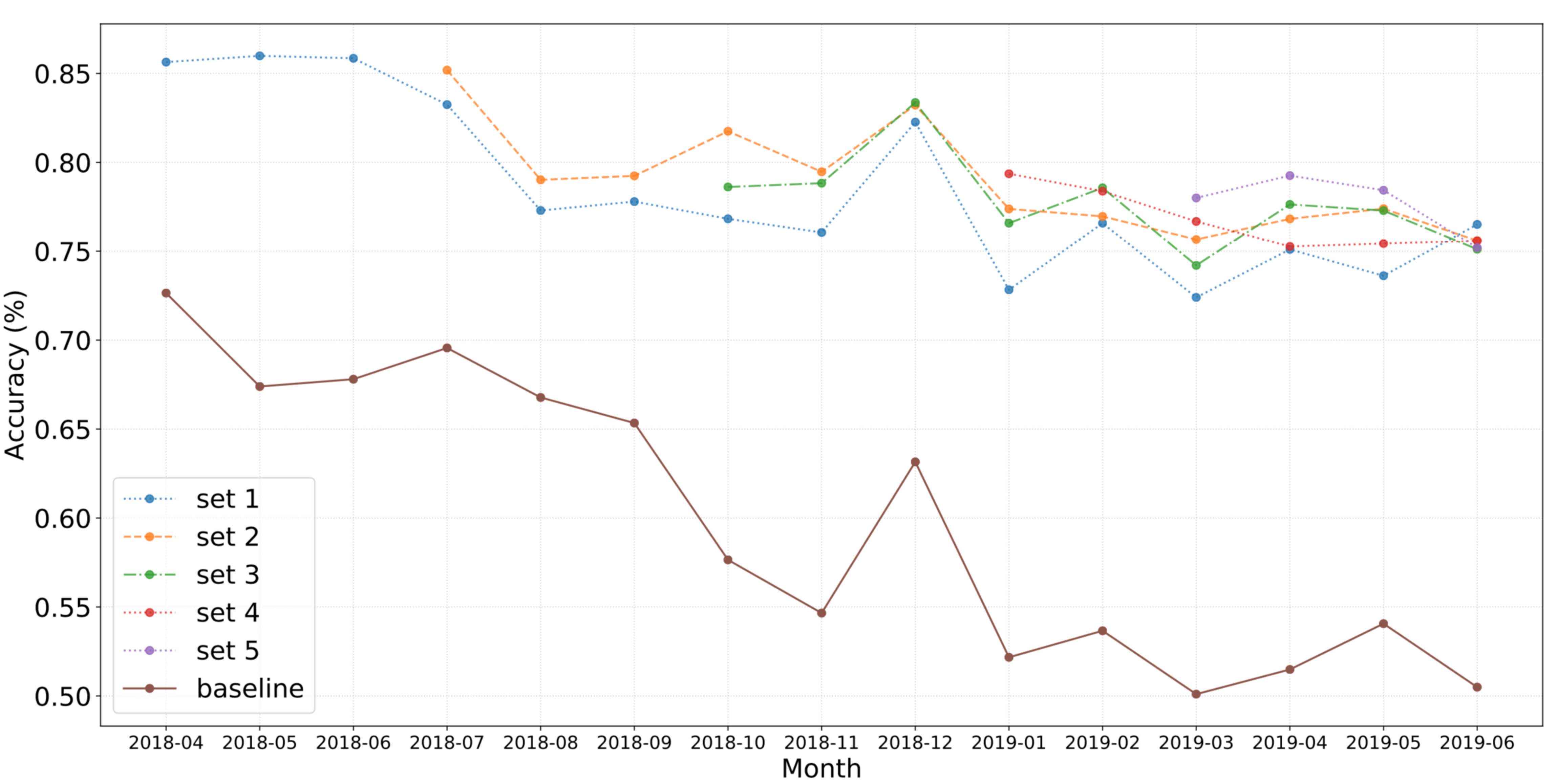}
    \caption{Monthly accuracy for each snapshot from Table \ref{tab:results2}. We also show the baseline, that in this case is the class Late. We can see that the snapshot 5 is the most recent and the one with better accuracy. The clients are performing better over the months since the proportion of invoices late and on time are decreasing over the months.}
    \label{fig:accuracysets}
\end{figure}

Despite the fact that in Table \ref{tab:results2} the accuracy was more or less the same we can see that in Figure \ref{fig:accuracysets} the last dataset, Set 5, has the best accuracy  monthly instead of globally. We can confirm using the baseline, that in this case is the class Late, that the proportion is decreasing over the years. Also, about seasonality we saw a spike at December and the same spike is reflected in the model. But even with this, the model keep some constant close to 77\%. So our conclusion on this experiments is that we don't need a large volume of historical data, which is good not just for the client (keep all the data) but also to the model that will be lighter. 

\begin{table*}[h]
\footnotesize
\centering
\begin{tabular}{|l|l|l|l|r|r|r|r|r|r|} \hline 
Dataset & Train & Validation & Test & $Train_R$ & $Validation_R$ & $Test_R$ & Baseline & Accuracy & F1 \\ \hline
Set 1 & 06-17 - 11-17 & 12-17 - 03-18 & 04-18 - 06-19 & 29\% & 21\% & 50\% & 61\% & 79\% & 77\% \\ \hline
Set 2 & 09-17 - 02-18 & 03-18 - 06-18 & 07-18 - 06-19 & 36\% & 22\% & 41\% & 57\% & 79\% & 78\% \\ \hline
Set 3 & 12-17 - 05-18 & 06-18 - 09-18 & 10-18 - 06-19 & 42\% & 23\% & 34\% & 53\% & 78\%  & 77\% \\ \hline
Set 4 & 03-18 - 08-18 & 09-18 - 12-18 & 01-19 - 06-19 & 49\% & 22\% & 28\% & 51\% & 77\% & 77\%\\ \hline
Set 5 & 06-18 - 11-18 & 12-18 - 02-19 & 03-19 - 06-19 & 55\% & 21\% & 24\% & 51\% & 78\%  & 78\% \\ \hline  
\end{tabular}
\caption{We generate 5 snapshots varying the time of training data and use the rest of data to validation and test. Here we present the snapshots information as period of training, period of validation and period of Test and also the training, validation and test ratios, the baseline and the global accuracy and F1 measure.}
\label{tab:results2}
\end{table*}

\begin{figure}[hbt]
    \centering
    \includegraphics[width=0.47\textwidth]{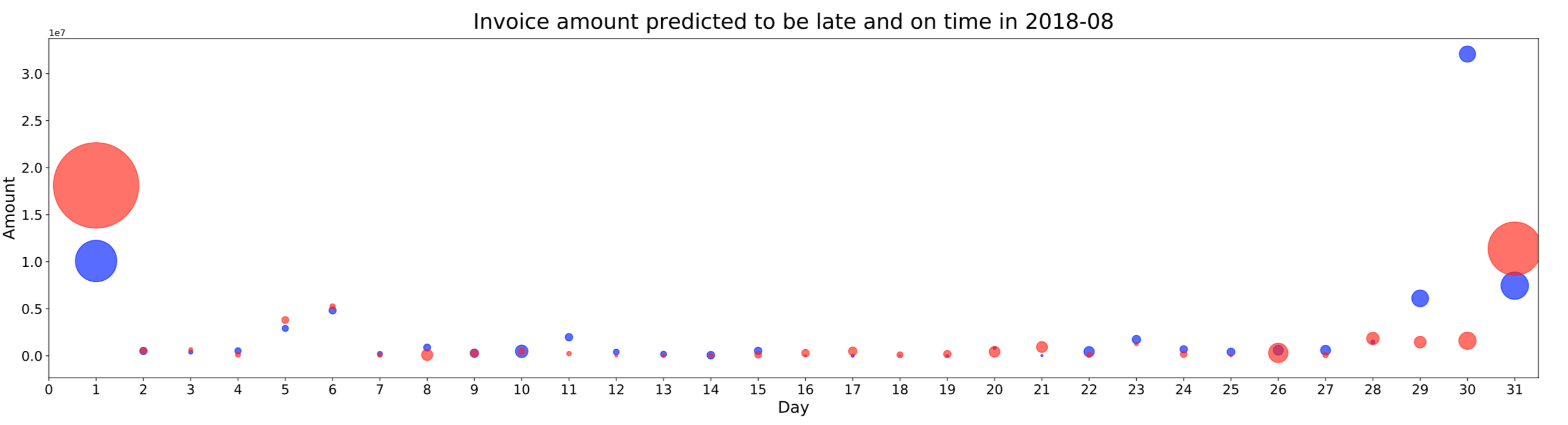}
    \caption{Invoices to be receive over one month, distributed over the payment data. Each point could have one or more invoices.}
    \label{fig:montlycollectpredict} 
\end{figure}

With the model trained we finally can use if to help collector optimize AR processes. Figure \ref{fig:montlycollectpredict} gives an example of this. For each month when a new invoice is created the model will give a prediction if the invoice will be payed Later or On Time, with this in hands, collectors can have a better vision of the process and focus in clients that are predict to be late and only follow the invoices that are predict to be on time to check if they were really payed. As we will show in the Section \ref{sec:ranking} we also proposed a new way to prioritizing the clients that before the use of machine learning were sorted mainly by the amount of money that they are in debt. Now, with the model we can use the probability of an invoice be payed late to adjust this sort. 

\section{Invoice Prioritization} \label{sec:ranking}

In the previous section we demonstrated that we can effectively predict the probability of an invoice being late. Identifying invoices that are likely to be delinquent at the time of creation enables us to steer the collections process, thus helping to save resources~\cite{zeng2008using}. However, our focus here is on taking actions by customers, not invoices. Furthermore, considering that resources are finite, we can not work with the hypothesis that all invoices are the same. In other words, we have to associate the invoice's probability of being late to its value (in dollars) so that we are able to measure the delinquency risk based not only on probabilities, but also total invoice's amount.

Currently, collectors' actions can be thought of as a greedy approach, that is, one that is solely based on an invoice's amount overdue. 
Which means that if an invoice $I_1$ has low probability of being late, let's say $P_{I_1}= 0.3506$ but a high value, $V_{I_1}=\$1,000,000.00$, in the usual ranking it will be in top position. On the other hand, if an invoice $I_2$ has a high probability of being late, $P_{I_2}= 0.9358$ but a low value, $V_{I_2}=\$300,000.00$ it will be in a low position in the usual ranking. 

As we show in Equation \ref{eq:rankingi}, we proposed to take into account its risk of being late, multiplying its probability by its value. In this way, we can continue prioritizing big customers and at the same time we save efforts on customers which will probably pay the invoice on time. 

\begin{equation}
\mathcal{R_{I_i}} = V_{I_i}*P_{I_i}( Y = Late )
\label{eq:rankingi}
\end{equation}

Next, we need to associate the invoice's level of information with a customer's level of information. We assumed that a client will be contacted about its total amount risk, rather than only about a single invoice. To create a rank by customer, we averaged invoices' risk by customer as shown in Equation \ref{eq:rankingc}.

\begin{equation}
\mathcal{R_{C_j}} = \frac{1}{N} \sum\limits_{i=1}^N \mathcal{R_{I_i}} 
\label{eq:rankingc}
\end{equation}

In order to compare our new prioritization ranking with older greedy approach used by collectors. We use Kendall's $\tau$ \cite{kendall1948rank} as a metric to compare the number of pairwise disagreements between two orders. Values close to 1 indicate strong agreement, values close to -1 indicates strong disagreement. Our new ranking has $\tau = 0.003$ which means that we change almost of $50\%$ of the ranking order. 

To conclude, our new list has a prioritization based on a more reliable metric rather than the greedy approach. 
This will benefit collectors by means of focusing their resources on reducing diligence, as it has been demonstrated that taking actions before an invoice being late can diminish collection time.

\section{Conclusion} \label{sec:conclusion}

In this paper, we built a model that computes the probability score of an invoice being overdue in the context of AR practices. This is critical when dealing with a very large set of invoices, which in turn requires collectors to rank customers and focus on those more likely to be delinquent. Our results are significant, with an accuracy of up to 77\%. The model developed in our work will be able to help our client attain a better sense of its AR operations and take better actions, thus improving its cash flow. 

Our set of historical features is small and captures the customer behavior payment using temporal information to make better prediction. We demonstrated by our experiments that using the window size with a small number of months (3) we were able to deal with concept drift in the dataset. We also created a new prioritization list that is able to rank customers in a more realistic way, helping the client to optimize their resources with respect to daily action of the collectors. 

Finally, as real world environments are always in continuous flux, our features distribution are shifting as well. We noted that the current AR process has been modified over the last year, and, as it evolves, our model should accompany that. We thus recommend a continuous evaluation of the analytics results so as to keep track of accuracy metrics as well as AUC, as discussed in the results section. From time to time, it seems necessary to retrain the model since all processes suffer from what is called concept drift, i.e., the relationship between the features and the labels evolve over time and classical machine learning approaches consider only stationary data.

Future work involves providing an additional microservice for building visual analytics components to support collectors in the task of identifying recent customers' behaviors. This need came out during interviews with collectors and SMEs. In these interviews, they mentioned to have some knowledge about customers' behaviors, for instance, that some clients always pay few days late or that a certain client is part of an industry domain facing financial challenges. Such visual analytics is planned to be delivered as part of the ranked list UI so that collectors can grasp the recent behavior of customers with respect to all paying activities of recent invoices. This should foster collectors contacting target clients by phone, for example. While a discussion and analysis of the interviews and the design of the visualization UI is out of the scope of this paper, the overall project sits in an activity that encompasses a broader sociotechnical arrangement, from the technical development (dealing with data processing, classification algorithms, and technology deployment) to the human practices (dealing with the understanding of the collectors activity, human perception when dealing with visualizations, and how to incorporate such technology as part of collectors daily activity).

To conclude, in this project we managed to delivery a machine learning model that successfully predicts the probability of an invoice of being late in order to rank customers and subsequently prioritize collectors' efforts by focusing on those more likely to be overdue. In doing so, this brings us closer to the data-driven paradigm of decision-making processes where companies rely on data to improve their activities and direct their business needs, thus saving resources and improving efficiency.

%

%
\bibliographystyle{plain}
\bibliography{igfbibliografy}

\begin{thebibliography}{10}

\bibitem{abe2010optimizing}
Naoki Abe, Prem Melville, Cezar Pendus, Chandan~K Reddy, David~L Jensen,
  Vince~P Thomas, James~J Bennett, Gary~F Anderson, Brent~R Cooley, Melissa
  Kowalczyk, et~al.
\newblock Optimizing debt collections using constrained reinforcement learning.
\newblock In {\em Proceedings of the 16th ACM SIGKDD international conference
  on Knowledge discovery and data mining}, pages 75--84. ACM, 2010.

\bibitem{baesens2005neural}
Bart Baesens, Tony Van~Gestel, Maria Stepanova, Dirk Van~den Poel, and Jan
  Vanthienen.
\newblock Neural network survival analysis for personal loan data.
\newblock {\em Journal of the Operational Research Society}, 56(9):1089--1098,
  2005.

\bibitem{bailey1999}
Butler B. Smith T. Swift T. Williamson J. Scherer W.~T. Bailey, D.~R.
\newblock Providian financial corporation: Collections strategy.
\newblock In {\em Systems Engineering Capstone Conference}. University of
  Virginia, 1999.

\bibitem{Breiman:2001}
Leo Breiman.
\newblock Random forests.
\newblock {\em Mach. Learn.}, 45(1):5--32, October 2001.

\bibitem{cao2009modelling}
Ricardo Cao, Juan~M Vilar, and Andr{\'e}s Devia.
\newblock Modelling consumer credit risk via survival analysis.
\newblock {\em SORT: statistics and operations research transactions},
  33(1):0003--30, 2009.

\bibitem{cheong2018customer}
Michelle~LF Cheong and Wen SHI.
\newblock Customer level predictive modeling for accounts receivable to reduce
  intervention actions.
\newblock 2018.

\bibitem{Dirick2017}
Lore Dirick, Gerda Claeskens, and Bart Baesens.
\newblock Time to default in credit scoring using survival analysis: a
  benchmark study.
\newblock {\em Journal of the Operational Research Society}, 68(6):652--665,
  Jun 2017.

\bibitem{FRIEDMAN2002367}
Jerome~H. Friedman.
\newblock Stochastic gradient boosting.
\newblock {\em Computational Statistics \& Data Analysis}, 38(4):367 -- 378,
  2002.
\newblock Nonlinear Methods and Data Mining.

\bibitem{James:2014}
Gareth James, Daniela Witten, Trevor Hastie, and Robert Tibshirani.
\newblock {\em An Introduction to Statistical Learning: With Applications in
  R}.
\newblock Springer Publishing Company, Incorporated, 2014.

\bibitem{kendall1948rank}
Maurice~George Kendall.
\newblock Rank correlation methods.
\newblock 1948.

\bibitem{Lee:2013}
Elisa~T. Lee and John~Wenyu Wang.
\newblock {\em Statistical Methods for Survival Data Analysis}.
\newblock Wiley Publishing, 4th edition, 2013.

\bibitem{rychnovsky2018survival}
Michal Rychnovsk{\`y} et~al.
\newblock Survival analysis as a tool for better probability of default
  prediction.
\newblock {\em Acta Oeconomica Pragensia}, 2018(1):34--46, 2018.

\bibitem{smirnov2016modelling}
Janika Smirnov et~al.
\newblock {\em Modelling late invoice payment times using survival analysis and
  random forests techniques}.
\newblock PhD thesis, 2016.

\bibitem{tater2018prediction}
Tarun Tater, Sampath Dechu, Senthil Mani, and Chandresh Maurya.
\newblock Prediction of invoice payment status in account payable business
  process.
\newblock In {\em International Conference on Service-Oriented Computing},
  pages 165--180. Springer, 2018.

\bibitem{younes2013framework}
Bashar Younes.
\newblock {\em A Framework for Invoice Management in Construction}.
\newblock PhD thesis, University of Alberta, 2013.

\bibitem{zeng2008using}
Sai Zeng, Prem Melville, Christian~A Lang, Ioana Boier-Martin, and Conrad
  Murphy.
\newblock Using predictive analysis to improve invoice-to-cash collection.
\newblock In {\em Proceedings of the 14th ACM SIGKDD international conference
  on Knowledge discovery and data mining}, pages 1043--1050. ACM, 2008.

\end{thebibliography}

%


\end{document}